\documentclass{article}

\PassOptionsToPackage{numbers, sort}{natbib}

\usepackage[main, final]{neurips_2026}

\usepackage[utf8]{inputenc} % allow utf-8 input
\usepackage[T1]{fontenc}    % use 8-bit T1 fonts

\usepackage{url}            % simple URL typesetting
\usepackage{booktabs}       % professional-quality tables
\usepackage{amsfonts}       % blackboard math symbols
\usepackage{nicefrac}       % compact symbols for 1/2, etc.
\usepackage{microtype}      % microtypography
\usepackage{xcolor}         % colors
\usepackage{amsmath,amsfonts,bm}

\def\eqref#1{equation~\ref{#1}}
\def\1{\bm{1}}

\DeclareMathAlphabet{\mathsfit}{\encodingdefault}{\sfdefault}{m}{sl}
\SetMathAlphabet{\mathsfit}{bold}{\encodingdefault}{\sfdefault}{bx}{n}

\usepackage{graphicx}
\usepackage{subcaption}
\usepackage{caption}
\usepackage{duckuments}
\usepackage{multirow}
\usepackage{makecell}
\usepackage{enumitem}
\usepackage{colortbl}
\usepackage{xspace}
\usepackage{lipsum}
\usepackage{amssymb}
\usepackage{mathtools}
\usepackage{amsthm}
\usepackage{bbm}
\usepackage{listings}
\usepackage{wrapfig}
\usepackage{float}
\usepackage{minted}

\definecolor{RowHighlight}{gray}{0.9}

\usepackage{algorithm}
\usepackage{algpseudocode}

\theoremstyle{plain}

\theoremstyle{definition}

\theoremstyle{remark}

\usepackage[textsize=tiny]{todonotes}

\definecolor{refblue}{rgb}{0.21,0.49,0.74}

\usepackage[pagebackref,breaklinks,colorlinks,allcolors=refblue]{hyperref}
\usepackage[capitalize,noabbrev]{cleveref}

\newcommand{\sname}{LARC\xspace}

\usepackage{lipsum}
\usepackage{duckuments}

\newcommand*{\ShowNotes}{}

\ifdefined\ShowNotes
  \newcommand{\colornote}[3]{{\color{#1}\bf{#2: #3}\normalfont}}
\else
  \newcommand{\colornote}[3]{}
\fi

\title{Latent Reasoning in Continuous Space\\ for Unified Multimodal Models}

\author{
  Byungwoo Jeon$^{1}$ \; Yoonwoo Jeong$^{2}$ \; Hyunseok Lee$^{1}$ \; Minsu Cho$^{2,3}$\footnotemark[1] \; Jinwoo Shin$^{1,3}$\footnotemark[1] \\
  $^{1}$KAIST ~\quad $^{2}$POSTECH ~\quad $^{3}$RLWRLD  \\ \texttt{imbw2024@kaist.ac.kr} \\[5pt]
  \tt\ \small \textcolor{blue!60!black}{\href{https://rootyjeon.github.io/larc-umm/}{https://rootyjeon.github.io/larc-umm}}
  \vspace{-10pt}
}

\begin{document}

\begin{NoHyper}
\def\thefootnote{*}\footnotetext{Equal advising}
\end{NoHyper}
\def\thefootnote{\arabic{footnote}}

\maketitle
\begin{abstract}
Unified Multimodal Models (UMMs) achieve remarkable success on diverse multimodal understanding and generation tasks, yet they still struggle to effectively leverage language-based chain-of-thought reasoning for image generation.
As a result, existing reasoning-augmented UMMs often fail to inherit the test-time scaling and fine-grained controllability observed in large language models.
We attribute this failure to a fundamental \textit{modality mismatch}: reasoning is performed in a discrete token space, whereas image generation operates in a continuous space.
To bridge this gap, we introduce a Latent Reasoning framework in Continuous space (\sname), which alleviates the token-level bottleneck by reasoning over continuous hidden states.
Specifically, the training framework consists of two stages: (i) curriculum supervised fine-tuning (SFT), and (ii) information-gain reinforcement learning (RL).
First, the curriculum SFT stage gradually converts token-level reasoning into latent reasoning.
Subsequently, the information-gain RL stage self-evolves latent reasoning without ground truth supervision. 
Experiments across diverse text-to-image generation benchmarks show that \sname consistently improves generation quality, prompt following, and compositional fidelity.
Notably, \sname achieves state-of-the-art performance on GenEval with a 7\%p gain over BAGEL and demonstrates clear test-time scaling behavior, outperforming token-based generation under both parallel and sequential test-time scaling strategies.
\end{abstract}
\section{Introduction}
\label{sec:intro}

Recent advances in generative models have reshaped visual content creation, achieving high quality and fine-grained controllability in text-to-image synthesis, image editing, and instruction-following generation~\citep{rombach2022high, ramesh2022hierarchical, saharia2022photorealistic, ramesh2021zero, sun2024autoregressive}.
However, generation-only models still struggle with complex semantics that require language-level planning, such as attribute binding~\citep{chefer2023attend, feng2022training, rassin2023linguistic} or world modeling~\citep{ha2018world, hafner2019dream}.
Within this landscape, unified multimodal models (UMMs)~\citep{wu2025janus, xie2025showo2, qin2025uni, wang2024emu3, deng2025bagel}, have emerged as a compelling paradigm which integrates an image-generation module into large language models (LLMs) that contain rich textual semantics.

\begin{figure}[!tb]
    \centering
    \includegraphics[width=\linewidth]{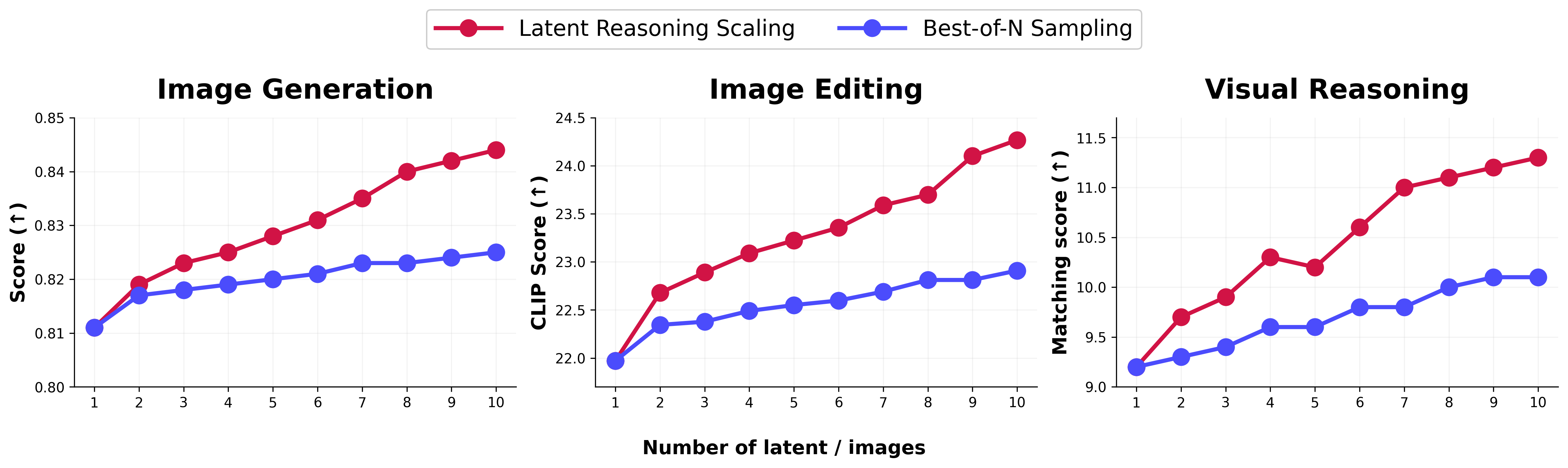}
    \caption{\textbf{Test-time scaling.} We investigate the effect of the number of latent during test-time. Under controlled experiments, \sname performs remarkable \textbf{test-time scaling behavior}, outperforming the best-of-N sampling on image generation~\citep{chang2025oneig}, image editing~\citep{jia2025compbench}, and visual reasoning~\citep{zhou2025mira} tasks.}
    \label{fig:2}
\end{figure}

Despite their success, prior UMMs fail to scale at test time: unlike in LLMs, increasing the reasoning budget does not yield the smooth performance gains observed on LLM benchmarks (Figure~\ref{fig:2}).
We argue that this failure stems from a fundamental \emph{modality mismatch}: reasoning is performed in a \emph{discrete} token space, whereas image generation operates in a \emph{continuous} latent space.
Since discrete chains struggle to encode the dense and spatially-grounded information that image synthesis demands~\citep{li2024autoregressive, fan2025fluid}, reasoning collapses into surface-level prompt manipulation, which is precisely the narrow regime in which existing methods provide gains.
This observation motivates our central claim: to extend the benefits of reasoning \emph{beyond} this regime, reasoning itself must operate in a continuous space.

To this end, we introduce \sname (\textbf{La}tent \textbf{R}easoning in \textbf{C}ontinuous space), a framework that mitigates the \textit{modality gap} by aligning text and image reasoning in a continuous latent space. 
Building on Coconut~\citep{hao2024training}, which performs latent reasoning in LLMs without explicit tokens, we extend the idea to multimodal generation. 
While we introduce reinforcement learning (RL) on the latent reasoning system, RL training faces two challenges: 
(i) Coconut-style reasoning is deterministic, whereas policy-gradient methods require stochastic rollouts; and 
(ii) latent reasoning traces lack an explicit evaluator, making it difficult to assess whether they are useful for generation.
We address both challenges with a two-stage recipe. 
Stage~1 employs curriculum supervised fine-tuning that progressively replaces token-level reasoning with continuous hidden states, thereby injecting controlled stochasticity and restoring the rollout diversity required for policy gradients. 
Then, Stage~2 applies reinforcement learning driven by an \emph{information-gain reward}, which quantifies how much each latent step improves the predictive distribution over the target image, together with a \emph{collapse-prevention reward} that preserves trajectory diversity. 
Together, these constitute a self-evolving loop that requires neither ground-truth chains nor external evaluators.

In experiments, we instantiate \sname on BAGEL~\citep{deng2025bagel} with 1M interleaved samples drawn from FLUX-Reason~\citep{fang2025flux}, WEAVE~\citep{weave2024}, and ScaleEdit~\citep{chen2026scaleedit}. 
On three text-to-image benchmarks---GenEval~\citep{ghosh2023geneval}, DPG-Bench~\citep{hu2024ella}, and OneIG-Bench~\citep{chang2025oneig}---\sname surpasses both the base UMMs and recent reasoning-augmented variants by substantial margins, \textit{e.g.}, $+7.6$ absolute points on OneIG-Bench over Uni-CoT~\citep{qin2025uni}.
More importantly, to our knowledge, \sname is the first reasoning-augmented UMM that exhibits clear test-time scaling: under budget forcing, doubling the latent reasoning compute yields monotonic gains across all three benchmarks.

Our contributions are summarized as follows:
\begin{itemize}[itemsep=0.01in,leftmargin=0.2in,topsep=0.02in, nosep]
    \item We identify a discrete--continuous \emph{modality mismatch} between reasoning and generation as the root cause of the limited gains and absent test-time scaling in current reasoning-augmented UMMs, and argue that reasoning must operate in a continuous space.
    \item We propose \sname, the first framework to perform reasoning in the same continuous latent space as image generation within a unified multimodal model, enabling test-time scaling in UMMs.
    \item We design a self-evolving training recipe that (i) extends Coconut-style latent reasoning with controlled stochasticity to make it amenable to RL, and (ii) introduces an \emph{information-gain} reward coupled with a \emph{collapse-prevention} reward, eliminating the need for ground-truth latent chains or an external critic.
\end{itemize}

\section{\sname: Latent Reasoning in Continuous Space} \label{sec:method}
\subsection{Overview}
We propose \sname, a framework that enables UMMs to perform image generation through latent reasoning on continuous space. We first describe the two-stage training pipeline of \sname. We then elaborate on the design of each component in the following sections. Specifically, in \Cref{sec:method:3.2}, we first define the interleaved trajectory format. Then, in \Cref{sec:method:3.3}, we introduce the curriculum supervised fine-tuning. Finally, in \Cref{sec:method:3.4}, we present reinforcement learning using information-gain reward derived from conditional mutual information.

\begin{figure}[t]
    \centering
    \includegraphics[width=\linewidth]{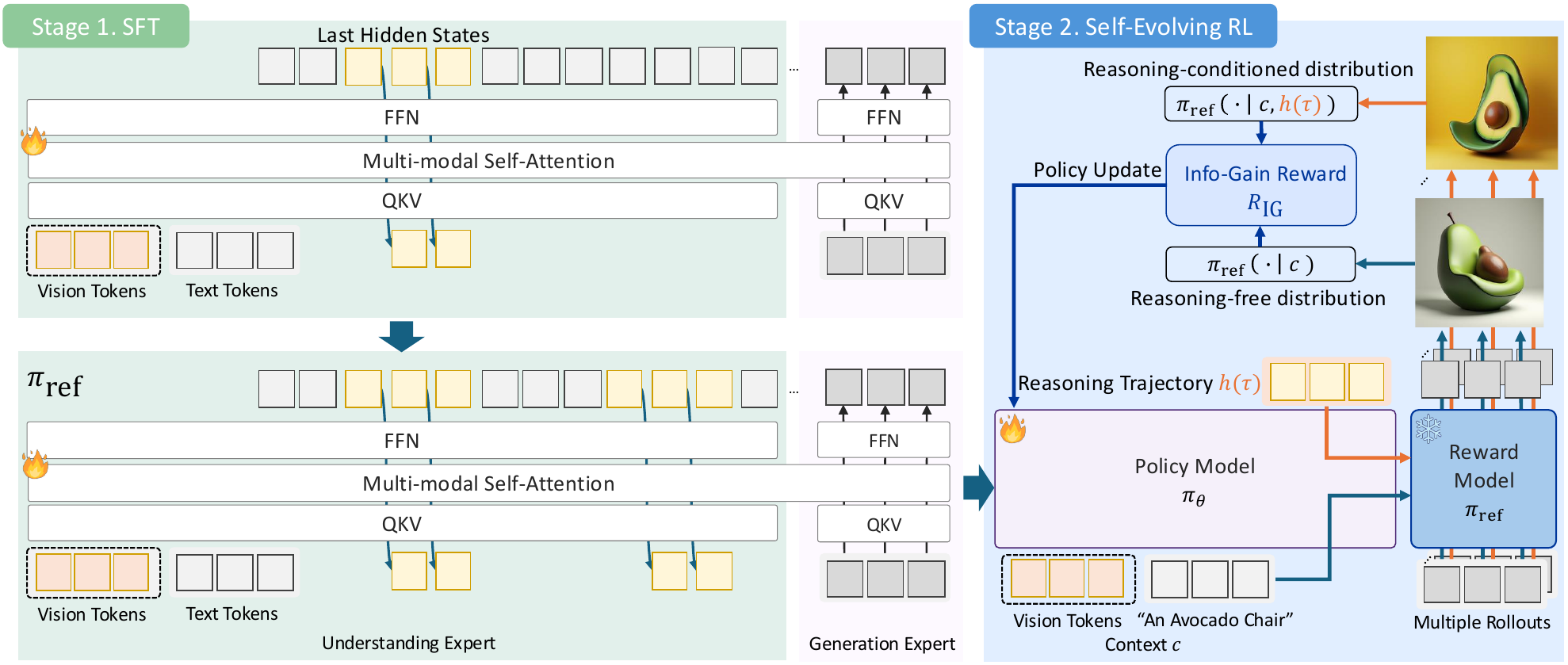}
    \caption{\textbf{Overview of \sname.} Our framework, \sname, consists of two stages. (1) \textbf{Stage 1} adapts supervised fine-tuning (SFT) while gradually introducing latent thought on continuous space where the last hidden state of UMM becomes the next input embedding of the model. (2) \textbf{Stage 2} adapts self-evolving reinforcement learning (RL) with information-gain reward derived from the internal knowledge. $\pi_{\text{ref}}$ denotes SFT model trained in Stage 1. Dotted line is task-specific option.}
    \label{fig:1}
\end{figure}

\subsection{Interleaved Latent Reasoning}\label{sec:method:3.2}
To enable continuous-space reasoning compatible with policy-gradient optimization, we formulate the reasoning trajectory as an alternating sequence of text segments and continuous latent steps, both generated by the UMM backbone.
Formally, given an input context $\mathbf{c}=(c_1 \cdots c_T)$ consisting of a text prompt and optionally a source image, the model produces a reasoning trajectory followed by a target image $\mathbf{x}$.
Following Coconut~\citep{hao2024training} and subsequent latent reasoning work~\citep{jeon2026vision}, the model alternates between two modes during trajectory generation: the \emph{language mode} and the \emph{latent mode}.
In language mode, the model emits text tokens through stochastic sampling, exposing the trajectory to the gradient signal of policy optimization.
In the latent mode, the model emits a continuous reasoning step that bypasses text decoding, preserving the rich semantic content of internal representations.

Specifically, the UMM backbone first processes the input embedding $E_T = [e(c_1), \cdots, e(c_T)]$ to produce the last hidden state $H_T = \text{Transformer}(E_T)$, where $e$ is the token embedding function. 
Once the model produces the special token \texttt{$<$latent$>$} at position $t$, the mode turns latent and the embedding of the next position is taken to be $h_t$ the hidden state at position $t$ in the last transformer layer. 
The mode returns to language after the fixed steps, at which point the trajectory ends and the image module is invoked. 
Concretely, with $E_t$ denoting the input embedding sequence up to position $t$ and $h_t = H_{t}[t, :]$ the corresponding hidden state, the next-position embeddings is 
\begin{align*}
    E_{t+1} &= 
    \begin{cases}
        [E_t; h_t] & \text{if latent mode,} \\
        [E_t; e(x_{t+1})] & \text{if language mode,}
    \end{cases} \\
    H_{t+1} &= \text{Transformer}(E_{t+1}).
\end{align*}

At inference, the mode alternation proceeds until the model emits the \texttt{<eos>} token, with each \texttt{<latent>} token marking the transition from a language-mode span to the next latent-mode step.
After termination, the full trajectory of hidden states $\mathbf{h}(\tau)$ is read by the image module through cross-attention to produce the target image $\mathbf{x}$.
The resulting reasoning trajectory naturally takes the form
\[
\tau = (s_1, l_1, s_2, l_2, \cdots, s_N, l_N),
\]
where $s_n$ denotes the $n$-th language-mode span and $l_n$ denotes the $n$-th latent-mode hidden state, each consisting of multiple tokens.
Here, the trajectory length $N$ and the length of each language-mode span $|s_n|$ are determined by when the model issues \texttt{<eos>} and \texttt{<latent>}, respectively, rather than fixed by an external schedule.

\begin{table}[t]
  \centering
  \caption{\textbf{\sname achieves strong text-to-image generation performance on GenEval.}
  We report accuracy (\%) on GenEval subcategories and the overall score. Bold and underline indicate the best and second-best results within each group, respectively.}
  \label{tab:geneval}
  \vspace{0.5em}
  \footnotesize
  \resizebox{\linewidth}{!}{
  \begin{tabular}{llccccccc}
    \toprule
    Type & Method 
    & Single Obj. $\uparrow$ 
    & Two Obj. $\uparrow$ 
    & Counting $\uparrow$ 
    & Colors $\uparrow$ 
    & Position $\uparrow$ 
    & Attr. Binding $\uparrow$ 
    & Overall $\uparrow$ \\
    \midrule
    \multirow{3}{*}{Reference}
    & GPT-4o~\citep{hurst2024gpt4o} 
    & 0.99 & 0.92 & \underline{0.85} & \underline{0.92} & 0.75 & 0.61 & \underline{0.84} \\
    & SD3-medium~\citep{esser2024stablediffusion3}
    & 0.99 & 0.94 & 0.72 & 0.89 & 0.33 & 0.46 & 0.74 \\
    & FLUX.1-dev~\citep{labs2025flux1kontextflowmatching}
    & 0.99 & 0.81 & 0.79 & 0.74 & 0.20 & 0.47 & 0.67 \\
    \midrule
    \multirow{8}{*}{UMM}
    & \multicolumn{8}{c}{Discrete UMM} \\
    & Janus-Pro~\citep{chen2025januspro}
    & 0.99 & 0.89 & 0.59 & 0.90 & \textbf{0.79} & 0.66 & 0.80 \\
    & Emu3~\citep{wang2024emu3}
    & 0.98 & 0.71 & 0.34 & 0.81 & 0.17 & 0.21 & 0.54 \\
    \cmidrule{2-9}
    & \multicolumn{8}{c}{Continuous UMM} \\
    & Show-o2~\citep{xie2025showo2}
    & \textbf{1.00} & 0.87 & 0.58 & \underline{0.92} & 0.52 & 0.62 & 0.76 \\
    & BAGEL~\citep{deng2025bagel}
    & 0.99 & 0.94 & 0.80 & 0.87 & 0.64 & 0.63 & 0.81 \\
    & Uni-CoT~\citep{qin2025uni}
    & 0.99 & \underline{0.96} & 0.84 & \underline{0.92} & 0.57 & \underline{0.71} & 0.83 \\
    \rowcolor{RowHighlight}& \sname (ours)
    & \textbf{1.00} & \textbf{0.98} & \textbf{0.86} & \textbf{0.93} & \underline{0.77} & \textbf{0.78} & \textbf{0.88} \\
    \bottomrule
  \end{tabular}
  }
  \vspace{-0.1in}
\end{table}

\subsection{Stage 1: Supervised Fine-Tuning with Curriculum Learning}\label{sec:method:3.3}
As observed in prior work~\citep{hao2024training}, introducing all $N$ latent steps from the beginning of training is unstable, since the latent stream lacks any direct supervision while the model has yet to learn the basic reasoning format.
To progressively guide the model toward reasoning in latent space, we perform supervised fine-tuning (SFT) under a curriculum that gradually converts token-level reasoning into latent reasoning.
We therefore decompose this stage into a curriculum that activates latent steps progressively, allowing each newly introduced latent step to build upon the trained prefix.

\noindent{\bf Data construction.}
For SFT with a curriculum, we collect 1M interleaved chain-of-thought (CoT) samples, providing the input context $\mathbf{c}$, a step-by-step reasoning chain, and the target image $\mathbf{x}$. We process each reasoning chain by inserting the special token \texttt{$<$latent$>$} at its step boundaries, marking the positions where latent steps are introduced.

\noindent{\bf Curriculum learning.}
To stabilize training, we adopt a curriculum learning that exposes the model to deeper interleaving over phase~\citep{hao2024training}. We first suppress all latent steps and train on the text-only trajectory $\tau^{(0)} = (s_1, \cdots, s_N)$, establishing the reasoning format under standard next-token prediction. Subsequently, in phase $k$ for $k = 1, \cdots, N-1$, we activate only the first $k$ latent steps, ensuring that each newly introduced latent is built on top of the stable prefix. The final phase activates all latent steps, with the learning rate reduced at each phase transition.

\noindent{\bf Training loss.}
We optimize the model $\pi_{\theta}$ with a objective that combines a standard next-token prediction loss $\mathcal{L}_{\text{SFT}}(\cdot)$ over text segments and a flow-matching loss $\mathcal{L}_{\text{flow}}(\cdot)$ on the target image, 
\[
\mathcal{L}_{\text{SFT}}^{(k)} = - \sum_{n=1}^{N} \log \pi_{\theta}\left(s_n | \mathbf{c}, s_{<n}, l_{<\min(n,k+1)}\right) + \mathcal{L}_{\text{flow}} \left( \mathbf{x} | \mathbf{c}, s_{1:N}, l_{1:k} \right),
\]
where $s_{<n}$ and $l_{<\min(n,k+1)}$ denote the text segments and the active latent steps preceding $s_n$ at phase $k$, respectively. 
Note that we propagate the gradient through the latent steps without explicit supervision target. This implicit supervision allows the latent steps to encode useful representations automatically, imposing no external bias. 

\subsection{Stage 2: Self-Evolving Reinforcement Learning}\label{sec:method:3.4}

Building on the latent reasoning capability established in Stage 1, we further refine the latent of reasoning trace through reinforcement learning (RL), guiding the model toward latent thoughts that are demonstrably useful for image generation. 
While supervised fine-tuning (SFT) effectively guides the model to produce interleaved trajectories, it remains fundamentally limited at refining the latent as it depends highly on the data quality, even the reasoning chain does not have ground-truth latent annotations. 
In this stage, we therefore adapt self-rewarding RL with an intrinsic reward derived from the UMM itself, yielding a fully self-evolving framework without any external critic.

\begin{table}[t]
  \centering
  \caption{\textbf{\sname improves prompt-following fidelity on DPG-Bench.}
  We report scores across global, entity-level, attribute, relation, and other categories. Higher is better.}
  \label{tab:dpg_bench}
  \vspace{0.5em}
  \footnotesize
  \begin{tabular}{llcccccc}
    \toprule
    Type & Method 
    & Global $\uparrow$ 
    & Entity $\uparrow$ 
    & Attribute $\uparrow$ 
    & Relation $\uparrow$ 
    & Other $\uparrow$ 
    & Overall $\uparrow$ \\
    \midrule
    \multirow{2}{*}{Reference}
    & SD3-medium~\citep{esser2024stablediffusion3}
    & 87.90 & 91.01 & 88.83 & 80.70 & 88.68 & 84.08 \\
    & FLUX.1-dev~\citep{labs2025flux1kontextflowmatching}
    & 82.10 & 89.50 & 88.70 & 91.10 & 89.40 & 84.00 \\
    \midrule
    \multirow{8}{*}{UMM}
    & \multicolumn{7}{c}{Discrete UMM} \\
    & Janus-Pro~\citep{chen2025januspro} 
    & 86.90 & 88.90 & 89.40 & 89.32 & 89.48 & 84.19 \\
    & Emu3~\citep{wang2024emu3}
    & 85.21 & 86.68 & 86.84 & 90.22 & 83.15 & 80.60 \\
    \cmidrule{2-8}
    & \multicolumn{7}{c}{Continuous UMM} \\
    & Show-o2~\citep{xie2025showo2}
    & \underline{89.00} & \underline{91.78} & 89.96 & \underline{91.81} & \underline{91.64} & \underline{86.14} \\
    & BAGEL~\citep{deng2025bagel}
    & 88.94 & 90.37 & 91.29 & 90.82 & 88.67 & 85.07 \\
    & Uni-CoT~\citep{qin2025uni}
    & 87.58 & 89.14 & \underline{91.62} & 91.09 & 86.97 & 85.56 \\
    \rowcolor{RowHighlight}& \sname (ours)
     & \textbf{92.25} & \textbf{91.89} & \textbf{93.00} & \textbf{94.70} & \textbf{92.35} & \textbf{90.04} \\
    \bottomrule
  \end{tabular}
\end{table}
\begin{table}[t]
  \centering
  \caption{\textbf{\sname substantially improves instruction-guided image generation on OneIG-Bench.}
  We report performance across prompt types and prompt lengths. NP refers natural language prompt. T\&P denotes tag-based and phrase-based prompt. Higher is better.}
  \label{tab:oneig_bench}
  \vspace{0.5em}
  \footnotesize
  \begin{tabular}{lcccccc}
    \toprule
    Method 
    & NP $\uparrow$ 
    & T\&P $\uparrow$ 
    & Short $\uparrow$ 
    & Medium $\uparrow$ 
    & Long $\uparrow$ 
    & Overall $\uparrow$ \\
    \midrule
    & \multicolumn{6}{c}{Discrete UMM} \\
    Janus-Pro~\citep{chen2025januspro}
    & 0.557 & 0.533 & 0.609 & 0.548 & 0.515 & 0.552 \\
    BLIP-o3~\citep{chen2025blipo3} 
    & 0.719 & 0.671 & 0.754 & 0.712 & 0.674 & 0.706 \\
    \midrule
    & \multicolumn{6}{c}{Continuous UMM} \\
    Show-o2~\citep{xie2025showo2}
    & \underline{0.825} & \underline{0.778} & \underline{0.825} & \underline{0.819} & \underline{0.807} & \underline{0.811} \\
    BAGEL~\citep{deng2025bagel}
    & 0.776 & 0.734 & 0.782 & 0.769 & 0.759 & 0.764 \\
    % BAGEL~\citep{deng2025bagel} + CoT 
    % & 0.798 & \underline{0.767} & \underline{0.824} & 0.793 & 0.767 & \underline{0.790} \\
    Uni-CoT~\citep{qin2025uni}
    & 0.803 & 0.748 & 0.817 & 0.805 & 0.776 & 0.790 \\
    \rowcolor{RowHighlight}\sname (ours) & \textbf{0.892} & \textbf{0.840} & \textbf{0.887} & \textbf{0.869} & \textbf{0.840} & \textbf{0.866} \\
    \bottomrule
  \end{tabular}
\end{table}

\noindent{\bf Information-Gain reward.} 
% Inspired by recent test-time scaling methods~\citep{tang2026thinking, jang2025verifier}, 
Motivated by prior work~\citep{jang2025verifier}, we define the reward as the \textit{Information Gain} that a reasoning trace provides to a frozen reference model, measuring how much the latent reasoning improves the prediction of the target image.
Concretely, letting $\pi_{\text{ref}}$ denote the model after Stage 1 and marginal density of its image module $p_{\pi_{\text{ref}}}(\mathbf{x}|\cdot)$, the reward for a sampled trajectory $\tau$ is 
\begin{equation}\label{eq1}
r_{\text{IG}}(\tau; \mathbf{x}, \mathbf{c}) \triangleq \log p_{\pi_{\text{ref}}} \left( \mathbf{x} | \mathbf{c}, h(\tau) \right) - \log p_{\pi_{\text{ref}}} \left( \mathbf{x} | \mathbf{c} \right),
\end{equation}

where $p_{\pi_{\text{ref}}} \left( \mathbf{x} | \mathbf{c}, h(\tau) \right)$ is the predictive density when the trace is included and $p_{\pi_{\text{ref}}} \left( \mathbf{x} | \mathbf{c} \right)$ is the reasoning-free baseline of the same model. Equation~\ref{eq1} assigns reward exactly when the trace raises likelihood of the target image in reference model, and zero reward when the trace carries no information beyond the context $c$.

Specifically, the Information-Gain reward $r_{\text{RG}}$ is the per-sample form of conditional mutual information, with $\mathbb{E}[r_{\text{IG}}] = I_{\pi_{\text{ref}}}\left( \mathbf{x}; h(\tau) | \mathbf{c} \right) \geq 0$ under the joint distribution. Maximizing $r_{\text{IG}}$ is therefore equivalent to maximizing the mutual information between the latent reasoning trace and the target image, with the bare context $c$ playing the role of the marginal baseline. However, since the image module of UMM does not provide direct access to log-likelihoods, we thereby estimate $r_{\text{IG}}$ as a paired difference of conditional flow-matching losses under shared noise,
\begin{equation}\label{eq:ig_estimator}
r_{\text{IG}}(\tau; \mathbf{x}, \mathbf{c}) \;\approx\; \mathcal{L}_{\text{flow}}^{\pi_{\text{ref}}}(\mathbf{x}, \mathbf{c}) \;-\; \mathcal{L}_{\text{flow}}^{\pi_{\text{ref}}}\!\left(\mathbf{x}, \mathbf{c}, h(\tau)\right),
\end{equation}
where $\mathcal{L}_{\text{flow}}^{\pi_{\text{ref}}}(\mathbf{x}, \xi)$ denotes the conditional flow-matching loss of the reference image module under conditioning $\xi$, with the two evaluations sharing the same noise sample for sample-wise constant cancellation. We provide the full derivation in Appendix~\ref{app:info_gain_deriv}.

\noindent{\bf Total reward.}
To prevent the latent collapse where all $N$ latent steps within a trajectory converge into identical vectors, we add an inter-step regularization for latent diversity,
\[
r_{\text{div}}(\tau) = \frac{1}{N(N-1)} \sum_{n \neq n'} \left( 1 - \cos(l_n, l_{n'} )\right),
\]

where $\cos(\cdot, \cdot)$ denotes the cosine similarity between two latent steps, $l_n$ is the $n$-th latent-mode hidden state in trajectory $\tau$, and $N$ is the number of latent steps. The total reward used for policy optimization is then
\begin{equation}\label{eq:total_reward}
R(\tau; \mathbf{x}, \mathbf{c}) \;=\; r_{\text{IG}}(\tau; \mathbf{x}, \mathbf{c}) \;+\; \lambda_{\text{div}} \cdot r_{\text{div}}(\tau).
\end{equation}

\noindent{\bf Policy optimization.}
We optimize the policy $\pi_\theta$ through reinforcement learning with the following objective:
\begin{equation}
\max_{\pi_{\theta}}\;
\mathbb{E}_{(\mathbf{c},\mathbf{x})\sim \mathcal{D}}\left[\mathbb{E}_{\tau\sim\pi_{\theta}(\cdot\mid\mathbf{c})}\left[ R(\tau; \mathbf{x}, \mathbf{c}) \right]\right].
\label{eq:rl_obj}
\end{equation}

We adopt GRPO~\citep{shao2024deepseekmath} as the optimization algorithm, which estimates advantages from group rollouts and removes an explicit value model. 
Its simplicity and stability allow a single set of hyperparameters to transfer across all benchmarks. 
For more optimization details, please see Appendix~\ref{app:implementation}.
\section{Experiments}
\label{sec:experiments}

We provide an empirical evaluation of \sname by investigating the following questions:
\begin{itemize}[leftmargin=*,topsep=0.0pt,itemsep=.5pt]
    \item \textbf{RQ1}: Does \sname improve the performance on various tasks, \textit{i.e.}, image generation, image editing, visual reasoning? (Table~\ref{tab:geneval}, \ref{tab:dpg_bench}, \ref{tab:oneig_bench}, \ref{tab:editing}, \ref{tab:understanding})
    \item \textbf{RQ2}: Does \sname enable test-time scaling of UMMs? (Figure~\ref{fig:2})
    \item \textbf{RQ3}: Which component mostly contributes to \sname? (Table~\ref{tab:ablation_reasoning}, \ref{tab:ablation_integration}, \ref{tab:ablation_stage})
\end{itemize}

\textbf{Evaluation Setup.}
We evaluate Unified Multimodal Models (UMMs) on various tasks including Text-to-Image (T2I) generation, image editing, and multimodal understanding. 
Specifically, we choose GenEval~\citep{ghosh2023geneval}, DPG-Bench~\citep{hu2024ella},
%, T2I-Compbench~\citep{huang2023t2icompbench}, 
and OneIG-Bench~\citep{chang2025oneig} for T2I benchmarks and CompBench~\citep{jia2025compbench} for image editing benchmark. For multimodal understanding, we choose MME~\citep{fu2023mme}, MMB~\citep{liu2024mmb}, MMMU~\citep{yue2024mmmu}, and MIRA~\citep{zhou2025mira}. For T2I generation and image editing tasks, we follow the evaluation setup of target benchmarks, which mainly reports model-based metrics, including MLLM-as-a-Judge. We provide detailed evaluation setups in Appendix~\ref{app:eval_details}.

\textbf{Training Setup.}
In the main experiments, we trained \sname on BAGEL~\citep{deng2025bagel}. 
We perform supervised fine-tuning and reinforcement learning on a fixed 1M scale of Chain-of-Thought (CoT) data. 
We construct the dataset sampled from open-source datasets, \textit{e.g.}, FLUX-Reason~\citep{fang2025flux}, WEAVE~\citep{weave2024}, ScaleEdit~\citep{chen2026scaleedit}. We provide more details in Appendix~\ref{app:training_details}.

\textbf{Baselines.}
We compare \sname with image generation models, \textit{e.g.}, Stable Diffusion3~\citep{esser2024stablediffusion3} and FLUX.1-dev~\citep{fang2025flux}, and foundation models in unified multimodal modeling, \textit{e.g.}, BAGEL~\citep{deng2025bagel}, Show-o2~\citep{xie2025showo2}, Emu3~\citep{wang2024emu3}, and Janus-Pro~\citep{chen2025januspro}, and reasoning-enhanced UMM, Uni-CoT~\citep{qin2025uni}.

\subsection{Image Generation Experiments}

We evaluate the performance on various Text-to-Image (T2I) and image-editing benchmarks. Here, we mainly compare \sname with other UMMs and generation models, reporting score for each subcategory and overall averages. 
Specifically, we evaluate our \sname in (i) GenEval~\citep{ghosh2023geneval}, (ii) DPG-Bench~\citep{hu2024ella}, (iii) OneIG-Bench~\citep{chang2025oneig}, and (iv) CompBench~\citep{jia2025compbench}.

\textbf{\sname improves text-to-image generation performance.}
As shown in Table~\ref{tab:geneval}, \sname outperforms or consistently matches competing baselines across the evaluated objectives, demonstrating broad gains in T2I performance.
The improvement is most pronounced on Attribute Binding, with a +15\%p increase over the baseline.
This subtask requires the joint handling of visual realization and compositional planning, suggesting that multimodal generation-aware latent reasoning is particularly effective for such compositionally demanding scenarios.

\textbf{\sname improves prompt-following fidelity.} 
We validate the prompt-following fidelity of \sname against competing baselines on DPG-Bench~\citep{hu2024ella} and OneIG-Bench~\citep{chang2025oneig}.
As shown in Table~\ref{tab:dpg_bench} and \ref{tab:oneig_bench}, \sname achieves consistent improvements over competing baselines.
On DPG-Bench, the overall score of \sname improves by +5.8\% ($85.07 \rightarrow 90.04$) over the base model BAGEL.
On OneIG-Bench, \sname yields a +10.2\%p gain over the same base model.
These improvements indicate that our latent reasoning is particularly effective in prompt-following generation.
The consistent gains across every subcategory further suggest that \sname generalizes broadly across diverse prompt types.

\begin{table}[t]
  \centering
  \caption{\textbf{\sname improves image editing on CompBench.} We report performance across score of CLIP~\citep{radford2021learning} for text (LC-T) and image (LC-I), PSNR, SSIM and LPIPS. LC-T refers local CLIP score between predicted foreground and descriptions while LC-I refers CLIP similarity between predicted foreground and ground-truth image. Higher is better except for LPIPS.}
  \label{tab:editing}
  \vspace{0.5em}
  \footnotesize
  \begin{tabular}{lccccc}
    \toprule
    Method 
    & LC-T $\uparrow$ 
    & LC-I $\uparrow$ 
    & PSNR $\uparrow$ 
    & SSIM $\uparrow$ 
    & LPIPS $\downarrow$ \\
    \midrule
    Janus-Pro~\citep{chen2025januspro}
    & 19.769 & 0.801 & 21.192 & 0.856 & 0.097 \\
    Emu3~\citep{wang2024emu3}
    & 19.536 & 0.803 & 22.036 & 0.868 & 0.092 \\
    Show-o2~\citep{xie2025showo2}
    & 20.350 & 0.831 & 24.661 & 0.919 & 0.065 \\
    BAGEL~\citep{deng2025bagel}
    & \underline{20.434} & \underline{0.842} & \underline{24.370} & \underline{0.917} & \underline{0.069} \\
    Uni-CoT~\citep{qin2025uni}
    & 20.215 & 0.837 & 23.925 & 0.904 & 0.088 \\
    \rowcolor{RowHighlight}\sname (ours)
    & \textbf{22.509} & \textbf{0.880} & \textbf{25.710} & \textbf{0.953} & \textbf{0.049} \\
    \bottomrule
  \end{tabular}
  \vspace{-0.2in}
\end{table}
\begin{table}[t]
  \centering
  \caption{\textbf{\sname consistently enhances multimodal understanding of UMM.} We report matching score for MME, MMB, MMMU, and MIRA.}
  \label{tab:understanding}
  \vspace{0.5em}
  \footnotesize
  \begin{tabular}{lcccc}
    \toprule
    Method 
    & MME~\citep{fu2023mme} $\uparrow$ 
    & MMB~\citep{liu2024mmb} $\uparrow$ 
    & MMMU~\citep{yue2024mmmu} $\uparrow$ 
    & MIRA~\citep{zhou2025mira} $\uparrow$ \\
    \midrule
    Janus-Pro~\citep{chen2025januspro}
    & 1567.1 & 79.2 & 41.0 & 4.9 \\
    Emu3~\citep{wang2024emu3}
    & 1487.3 & 67.4 & 31.6 & 4.1 \\
    Show-o2~\citep{xie2025showo2}
    & 1620.5 & 79.3 & 48.9 & 6.9 \\
    BAGEL~\citep{deng2025bagel}
    & 1687.0 & 85.0 & 55.3 & 7.5 \\
    Uni-CoT~\citep{qin2025uni}
    & \underline{1691.7} & \underline{85.4} & \underline{55.5} & \underline{10.6} \\
    \rowcolor{RowHighlight}\sname (ours)
    & \textbf{1725.3} & \textbf{88.1} & \textbf{56.2} & \textbf{11.7} \\
    \bottomrule
  \end{tabular}
\end{table}
\textbf{\sname improves image editing.}
As shown in Table~\ref{tab:editing}, \sname consistently outperforms all baselines across every metric.
\sname achieves a +10.2\% gain on LC-T over the base model BAGEL ($20.434 \rightarrow 22.509$), indicating substantially stronger adherence to editing instructions.
Pixel-level quality also improves markedly, with PSNR increasing from $24.3$ to $25.7$ and LPIPS decreasing from $0.069$ to $0.049$.
These results suggest that continuous latent reasoning enables the model to plan spatially grounded edits, particularly for localized attribute modification.

\subsection{Text Generation Experiments}

While the previous section established the benefits of latent reasoning for image generation and editing, an open question is whether such continuous reasoning also benefits text generation tasks that share the same UMM backbone.
We address this question by evaluating \sname on perception-heavy benchmarks (MME, MMB) and a visual reasoning benchmark (MIRA), comparing against the base model BAGEL and Uni-CoT, a recent UMM with discrete token-level reasoning.

As shown in Table~\ref{tab:understanding}, \sname improves over both BAGEL and Uni-CoT across all benchmarks.
The most pronounced gain appears on MIRA ($7.5 \rightarrow 11.7$), a benchmark requiring visual chain-of-thought reasoning that explicitly demands multi-step intermediate inference.
The substantial margin over Uni-CoT, which performs reasoning through discrete tokens, supports our hypothesis that latent reasoning provides richer intermediate representations than its discrete counterpart.
We attribute this to the fact that continuous latent steps preserve information that discretization through a fixed vocabulary inevitably collapses, an effect amplified in tasks where reasoning must integrate fine-grained visual cues.
Gains on MME and MMB further confirm that the continuous reasoning capability transfers positively to perception-heavy tasks without sacrificing general visual understanding.
This indicates that the additional latent reasoning supervision introduced during training does not degrade the original visual perception ability of the UMM backbone, an important property for unified models that serve both generation and understanding.

\subsection{Test-time Scaling}
As shown in Figure~\ref{fig:2}, we investigate the effect of latent reasoning at test-time. \sname exhibits monotonically increasing performance on image generation, image editing, and visual reasoning as the latent budget grows, a behavior that is absent in prior reasoning-augmented UMMs. Furthermore, \sname consistently outperforms best-of-N sampling under matched compute budgets across all three tasks. We attribute this scaling behavior to continuous latent reasoning operating in the same space as image generation, allowing additional compute to refine useful representations rather than producing redundant discrete tokens.

\subsection{Component Ablations}
\begin{table}[t]
  \centering
  \caption{\textbf{Effect of latent reasoning compared to token-level reasoning.} We report scores on GenEval for T2I generation, OneIG-Bench for compositional generation, CompBench for image editing, and MIRA for visual reasoning. Token-level Reasoning denotes standard Chain-of-Thought on discrete reasoning space. Latent Reasoning refers \sname in Stage 1 trained by SFT.}
  \label{tab:ablation_reasoning}
  \vspace{0.5em}
  \footnotesize
  \begin{tabular}{lccccc}
    \toprule
    Method 
    & Reasoning space
    & T2I gen. $\uparrow$ 
    & Comp. gen. $\uparrow$ 
    & Image editing $\uparrow$ 
    & Visual reasoning $\uparrow$ \\
    \midrule
    Baseline~\citep{deng2025bagel}
    & Discrete & 0.81 & 0.76 & 20.43 & 7.5 \\
    Token-level Reasoning
    & Discrete & 0.82 & 0.79 & 20.81 & 7.1 \\
    \rowcolor{RowHighlight}Latent Reasoning
    & Continuous & \textbf{0.83} & \textbf{0.81} & \textbf{21.46} & \textbf{9.7} \\
    \bottomrule
  \end{tabular}
  \vspace{-0.24in}
\end{table}
\begin{table}[t]
  \centering
  \caption{\textbf{Ablation on latent reasoning strategy.} We report scores on GenEval for T2I generation, OneIG-Bench for compositional generation, CompBench for image editing, and MIRA for visual reasoning. Naive refers latent reasoning without text segments proposed by the prior work~\citep{hao2024training}.}
  \label{tab:ablation_integration}
  \vspace{0.5em}
  \footnotesize
  \begin{tabular}{lcccc}
    \toprule
    Method 
    & T2I gen. $\uparrow$ 
    & Comp. gen. $\uparrow$ 
    & Image editing $\uparrow$ 
    & Visual reasoning $\uparrow$ \\
    \midrule
    Baseline~\citep{deng2025bagel}
    & 0.81 & 0.76 & 20.43 & 7.5 \\
    Naive
    & \textbf{0.83} & 0.79 & 21.02 & \textbf{9.9} \\
    \rowcolor{RowHighlight}Interleaved (ours)
    & \textbf{0.83} & \textbf{0.81} & \textbf{21.46} & 9.7 \\
    \bottomrule
  \end{tabular}
  \vspace{-0.24in}
\end{table}
\begin{table}[t]
  \centering
  \caption{\textbf{Ablation on training stages.} We report scores on GenEval for T2I generation, OneIG-Bench for compositional generation, CompBench for image editing, and MIRA for visual reasoning.}
  \label{tab:ablation_stage}
  \vspace{0.5em}
  \footnotesize
  \begin{tabular}{lcccc}
    \toprule
    Method 
    & T2I gen. $\uparrow$ 
    & Comp. gen. $\uparrow$ 
    & Image editing $\uparrow$ 
    & Visual reasoning $\uparrow$ \\
    \midrule
    Baseline~\citep{deng2025bagel}
    & 0.81 & 0.76 & 20.43 & 7.5 \\
    Stage 1
    & 0.83 & 0.81 & 21.46 & 9.7 \\
    \rowcolor{RowHighlight}Stage 1 \& 2
    & \textbf{0.88} & \textbf{0.87} & \textbf{22.51} & \textbf{11.7} \\
    \bottomrule
  \end{tabular}
  \vspace{-0.05in}
\end{table}

We conduct ablation studies to assess the contribution of each design choice in \sname, using GenEval, OneIG-Bench, CompBench, and MIRA as representative benchmarks.

\noindent{\bf Effect of reasoning space.} In Table~\ref{tab:ablation_reasoning}, We compare token-level reasoning and latent reasoning initialized from the same base model. Switching from token-level to latent reasoning yields consistent gains across all four benchmarks. The improvement is particularly pronounced on visual reasoning (+2.6 on MIRA) and image editing (+0.65 on LC-T), confirming that continuous representations better capture the dense, spatially-grounded information required for generation tasks. Notably, token-level reasoning slightly degrades visual reasoning performance compared to the baseline, suggesting that discrete reasoning suffers the modality gap.

\noindent{\bf Effect of latent reasoning strategy} In Table~\ref{tab:ablation_integration}, we compare two latent reasoning strategies: Naive latent reasoning without interleaved text segments and our proposed interleaved design. While the Naive approach provides reasonable gains, interleaved latent reasoning achieves better compositional generation (+0.02 on OneIG-Bench) and image editing (+0.44 on LC-T) with comparable performance elsewhere. We attribute this to the text segments maintaining stochastic rollout diversity required for policy gradient optimization, while the latent steps encode the continuous representations that image generation demands.

\noindent{\bf Ablation study for training stages.} In \Cref{tab:ablation_stage}, we conduct ablation study for each training stage. 
While Stage 1 (SFT with curriculum) yields consistent improvements over the baseline across all tasks, Stage 2 (self-evolving RL) further boosts performance substantially, with T2I generation improving from 0.83 to 0.88 and visual reasoning from 9.7 to 11.7. 
This confirms that Stage 2 successfully refines latent traces beyond what SFT can achieve, and validates the self-evolving design that requires neither ground-truth reasoning chains nor an external critic.
\section{Related Work}
\label{sec:related}

\subsection{Test-time Scaling in LLMs}

Test-time scaling is a model behavior where additional inference compute improves task performance in large language models (LLMs)~\citep{muennighoff2025s1,lee2025revise}.
Chain-of-Thought (CoT) prompting~\citep{wei2022chain} has emerged as a primary driver of test-time scaling in LLMs by decomposing problems into intermediate textual reasoning steps.
Subsequent work has internalized this capability through supervised fine-tuning~\citep{yue2023mammoth} and reinforcement learning~\citep{shao2024deepseekmath}.
Inference-time search methods further expand the reasoning space through tree-based~\citep{yao2023tree} and trajectory-based~\citep{gandhi2024stream} exploration.
Recognizing that natural language fails to encapsulate all forms of reasoning~\citep{lecun2022path}, recent paradigms shift reasoning directly into latent space~\citep{hao2024training, li2025latent, jeon2026vision}.
Despite these advances, prior efforts to transfer test-time scaling to UMM image generation~\citep{wang2024emu3, xie2025showo2, chen2025januspro, deng2025bagel, qin2025uni} have relied on discrete reasoning tokens and failed to exhibit the predictable scaling observed in language reasoning.
We attribute this failure to the modality mismatch between discrete reasoning tokens and continuous image generation.
Motivated by this observation, we propose \sname, which performs reasoning entirely in continuous space and becomes the first UMM to achieve test-time scaling for image generation.

\subsection{Image Generation Models}

Image generation has emerged as one of the most influential research directions in modern deep learning, advancing along two complementary paradigms.
Diffusion models~\citep{ho2020denoising, song2020score} have become the dominant paradigm for text-to-image generation, synthesizing images through iterative denoising in systems such as Stable Diffusion~\citep{rombach2022high}, DALL-E~2~\citep{ramesh2022hierarchical}, and Imagen~\citep{saharia2022photorealistic}.
In parallel, autoregressive approaches predict visual tokens from discrete vocabularies established by vector-quantized tokenizers~\citep{van2017neural, esser2021taming}, scaling from DALL-E~\citep{ramesh2021zero} to recent transformer-based models including LlamaGen~\citep{sun2024autoregressive} and VAR~\citep{tian2024visual}.
Building on these advances, unified multimodal models (UMMs) integrate visual understanding and generation within a single architecture, typically extending a pretrained LLM with image tokenization to inherit strong language priors and enable cross-task knowledge transfer~\citep{tong2025metamorph}.
Early work pursues fully discrete tokenization with next-token prediction: Chameleon~\citep{team2024chameleon} pioneers an early-fusion design trained from scratch, and Emu3~\citep{wang2024emu3} scales this paradigm across understanding and generation.
Hybrid formulations combine autoregressive text decoding with diffusion-based image generation: Show-o~\citep{xie2024show, xie2025showo2} uses discrete diffusion for images, and Transfusion~\citep{zhou2024transfusion} applies continuous diffusion over image patches.
A separate line of work decouples visual representations for the two tasks: Janus~\citep{wu2025janus} and Janus-Pro~\citep{chen2025januspro} introduce separate visual encoders under a shared transformer backbone, and BAGEL~\citep{deng2025bagel} extends decoupling to the transformer itself through a Mixture-of-Transformer-Experts architecture~\citep{liang2024mixture, shi2026lmfusion} combined with rectified flow~\citep{liu2022rectifiedflow} for image generation.
Our work is orthogonal to these architectural advances, introducing a reasoning capability that operates over an existing UMM backbone, with BAGEL being adopted as our base models.

\subsection{Reinforcement Learning in LLMs}

Reinforcement learning has long served as the standard machinery for policy optimization.
Trust-region methods such as TRPO~\citep{schulman2015trust} and its first-order successor PPO~\citep{schulman2017proximal} stabilize policy updates through constrained or clipped objectives.
More recent work adapts policy optimization to language modeling.
GRPO~\citep{shao2024deepseekmath} eliminates the value network by estimating advantages from group rollouts, and follow-up methods~\citep{liu2026gdpo, yu2025dapo, ma2026fipo} further improve this group-relative formulation.
These advances have powered substantial capability gains in LLMs, ranging from instruction-following in InstructGPT~\citep{ouyang2022instructgpt} to answer-verifiable reasoning in DeepSeek-R1~\citep{guo2025deepseekr1}, where GRPO drives chain-of-thought reasoning at scale.
However, for reasoning in image generation, no well-defined ground-truth reasoning trajectory exists.
This precludes direct supervision over the reasoning chain and motivates the RL-based formulation adopted in \sname.
We instantiate this formulation with GRPO, which is widely used in reinforcement learning tasks, to train reasoning chains using outcome-level rewards.
This bypasses the need for chain-level annotations standard in text-based CoT, yet absent in our setting.
\section{Conclusion}
\label{sec:con}

We introduced \sname, a latent reasoning framework that performs reasoning in continuous space for unified multimodal models (UMMs).
By representing reasoning steps as continuous hidden states rather than discrete tokens, \sname alleviates the modality gap that has prevented prior reasoning-augmented UMMs from inheriting the test-time scaling of large language models.
Our two-stage pipeline pairs curriculum supervised fine-tuning with a self-evolving reinforcement learning stage driven by an information-gain reward, requiring no external critic.
Instantiated on BAGEL with 1M interleaved samples, \sname achieves state-of-the-art text-to-image performance with a 7\%p gain over BAGEL and is the first reasoning-augmented UMM to exhibit clear test-time scaling for image generation.
This finding bridges a long-standing modality gap in UMMs and opens a path toward UMMs that match the inference-time scalability of modern reasoning LLMs.

\clearpage
\section*{Broader Impact Statement}

\sname advances the reasoning capability of unified multimodal models, with the goal of producing more controllable and inference-time-scalable image generation systems.
We discuss the broader impacts below.

\paragraph{Positive impacts.}
By enabling test-time scaling for image generation, \sname allows users to allocate additional inference compute to obtain higher-fidelity outputs, particularly for prompts that require fine-grained compositional reasoning such as attribute binding and spatial layout.
This controllability supports applications in design assistance, accessibility tools for visual communication, education, and scientific visualization, where prompt fidelity directly impacts utility.
The latent reasoning paradigm also offers a unified mechanism for transferring reasoning capability across modalities, potentially reducing the need for modality-specific reasoning pipelines and lowering the engineering cost of building multimodal systems.

\paragraph{Potential risks.}
As with other text-to-image generation systems, \sname inherits risks common to large generative models, including the potential for generating misleading or harmful imagery, amplifying biases present in pretraining data, and being misused for the production of deceptive content such as misinformation or non-consensual depictions.
The improved prompt-following fidelity demonstrated in our experiments increases generation quality across legitimate uses, and also amplifies these dual-use concerns, since outputs that more faithfully reflect adversarial prompts are correspondingly more capable of producing targeted harmful content.
We do not introduce new training data or new modalities beyond those already established in BAGEL and the public datasets we use, so risks specific to dataset provenance are inherited rather than newly created.

\paragraph{Mitigation and responsible deployment.}
We recommend that downstream applications of \sname combine the model with established safeguards, including content moderation classifiers, watermarking of generated images, prompt-level safety filters, and auditing of training data for licensing and bias.
For research deployments, releasing only model weights without bundled deployment infrastructure helps preserve the ability of downstream practitioners to apply context-appropriate safety measures.
We will release \sname under a license that prohibits use for generating disinformation, non-consensual intimate imagery, or content targeting protected groups, and we encourage the community to develop evaluation protocols specifically for reasoning-augmented generative models, where the relationship between reasoning fidelity and harmful output remains understudied.

\paragraph{Environmental considerations.}
Training \sname requires reinforcement learning over a multi-billion-parameter UMM backbone, which incurs non-trivial compute and associated carbon footprint.
We mitigate this by initializing from a pretrained BAGEL checkpoint rather than training from scratch, and by using a curriculum that converges in fewer steps than naive joint training of all latent reasoning steps.
Future work on parameter-efficient adaptation of latent reasoning, such as adapter-based or low-rank variants, would further reduce the resource cost of deploying this paradigm at scale.

\section*{Acknowledgments}
This work was partly supported by Institute for Information \& Communications Technology Planning \& Evaluation (IITP) grant funded by the Korea government (MSIT) (RS-2019-II190075, Artificial Intelligence Graduate School Program (KAIST); RS-2022-II220184, Development and Study of AI Technologies to Inexpensively Conform to Evolving Policy on Ethics; RS-2024-00509279, Global AI Frontier Lab), and RLWRLD Inc. This research was also supported by the NVIDIA Academic Grant Program using A100 GPUs on Brev.

The authors thank Suhyeok Jang, Junbeom Kim, and Seonghu Jeon for their insightful discussions and feedback on the experiments.

\clearpage

% --------------------------------------------------------------------
% References
% --------------------------------------------------------------------
{\small
\bibliography{ref}
\bibliographystyle{unsrtnat}
}

\newpage
\appendix
\appendix
\section{Implementation Details}
\label{app:implementation}

\subsection{Training Details}
\label{app:training_details}

We adopt BAGEL~\citep{deng2025bagel} as the base model and perform two-stage learning, namely (1) Supervised Fine-tuning (SFT) with curriculum and (2) self-evolving Reinforcement Learning (RL). In Stage 1, we freeze the vision tokenizer and jointly train the understanding expert and the generation expert of the UMM backbone. We follow the curriculum schedule described in Section~\ref{sec:method:3.3}, where the number of activated latent steps is increased from $0$ to $N$ over $N{+}1$ phases with the learning rate decayed at every phase transition. In Stage 2, we initialize the policy $\pi_\theta$ from $\mathcal{M}_{\text{SFT}}$ and freeze a copy of the same model as the reference $\pi_{\text{ref}}$. Only the understanding expert is updated during RL, since the generation expert is reused for reward estimation through the shared-noise flow-matching loss. Detailed hyperparameters are summarized in Table~\ref{tab:hyperparams}. All training experiments are conducted with 8$\times$ NVIDIA A100s, taking approximately 300 GPU hours in Stage 1 and 96 GPU hours in Stage 2.

\vspace{-0.18in}
\begin{table}[ht]
\centering
\caption{\textbf{Hyperparameters for Stage 1 and Stage 2.}}
\label{tab:hyperparams}
\small
\vspace{0.5em}
\begin{tabular}{lcc}
\toprule
Hyperparameter & Stage 1 (SFT) & Stage 2 (RL) \\
\midrule
Optimizer & AdamW & AdamW \\
Algorithm & - & GRPO \\
Learning rate & 1e-5 & 1e-6 \\
Global batch size & - & 64 \\
Max number of tokens & $6 \times 10^5$ & - \\
Weight decay & $10^{-2}$ & 0.0 \\
Training steps & $1.2 \times 10^3$ & $10^3$ \\
Group size $G$ & -- & 8 \\
Clipping $\epsilon_{\text{clip}}$ & -- & $2 \times 10^{-1}$ \\
KL coefficient $\beta$ & -- & $10^{-3}$ \\
Diversity weight $\lambda_{\text{div}}$ & -- & $10^{-1}$ \\
\bottomrule
\end{tabular}
\end{table}
\vspace{-0.18in}

\subsection{Evaluation Details}
\label{app:eval_details}

We evaluate \sname across three categories of benchmarks following the official protocol of each benchmark. For inference, we apply greedy sampling and generate one image per prompt unless specified otherwise.

\noindent\textbf{Text-to-Image generation.} We evaluate the models on GenEval~\citep{ghosh2023geneval}, DPG-Bench~\citep{hu2024ella}, and OneIG-Bench~\citep{chang2025oneig}. GenEval automatically verifies compositional properties of generated images via an object-detection pipeline, reporting accuracy across six sub-tasks, \textit{e.g.}, single object, two objects, counting, colors, position, and attribute binding. DPG-Bench evaluates 1{,}065 dense prompts through an MLLM-as-a-Judge protocol, providing scores across global, entity, attribute, relation, and other categories. OneIG-Bench spans multiple dimensions of text-to-image (T2I) evaluation, including prompt-image alignment, text rendering, and reasoning, with scores partitioned by prompt type (NP, T\&P) and prompt length (short, medium, long). Specifically, we report alignment score for OneIG-Bench.

\noindent\textbf{Image editing.} We evaluate the models on CompBench~\citep{jia2025compbench}, which features 3{,}200 complex instruction-guided editing tasks built through an MLLM-human collaborative framework. Specifically, we report multi-object editing tasks in Table~\ref{tab:editing}. Following the official protocol, we report local CLIP-Text (LC-T) score between the predicted foreground and the description, local CLIP-Image (LC-I) similarity between the predicted foreground and the ground-truth image, along with pixel-level metrics PSNR, SSIM, and LPIPS.

\noindent\textbf{Multimodal understanding.} We evaluate on MME~\citep{fu2023mme}, MMBench~\citep{liu2024mmb}, MMMU~\citep{yue2024mmmu}, and MIRA~\citep{zhou2025mira}. MME measures perception and cognition across 14 sub-tasks, while MMBench and MMMU benchmark broad multimodal reasoning across diverse disciplines. MIRA targets visual chain-of-thought reasoning with 546 carefully curated problems spanning 20 task types, where successful reasoning requires generating intermediate visual representations.

For API-based baselines, we report results using the model version \texttt{gpt-4o-2024-08-06} for GPT-4o~\citep{hurst2024gpt4o}. We adopt VLMEvalKit~\citep{duan2024vlmevalkit} as the unified evaluation toolkit for multimodal understanding tasks, and use the official codebase released by each benchmark for T2I generation and image editing evaluation.

\section{Data Statistics}
\label{app:data}

We construct a 1M scale reasoning dataset by sampling from three open-source datasets, namely FLUX-Reason~\citep{fang2025flux}, WEAVE~\citep{weave2024}, and ScaleEdit~\citep{chen2026scaleedit}. The detailed composition is summarized in Table~\ref{tab:data_stats}. Following the prior works~\citep{jeon2026spatialboost,jeon2026vision}, we balance the mixture so that text-to-image, image editing, and interleaved reasoning are jointly represented, ensuring that the latent reasoning trajectory generalizes across diverse generation tasks. For each sample, we insert the special token \texttt{<latent>} at every reasoning step boundary as described in Section~\ref{sec:method:3.3}, marking the positions where latent steps are progressively activated by the curriculum.

\vspace{-0.2in}
\begin{table}[ht]
\centering
\caption{\textbf{Composition of the 1M reasoning dataset.}}
\label{tab:data_stats}
\small
\vspace{0.5em}
\begin{tabular}{lcc}
\toprule
Source dataset & Task & \# samples \\
\midrule
FLUX-Reason~\citep{fang2025flux} & Text-to-Image & 500K \\
WEAVE~\citep{weave2024} & Text-to-Image & 100K \\
ScaleEdit~\citep{chen2026scaleedit} & Image Editing & 400K \\
\midrule
Total & -- & 1M \\
\bottomrule
\end{tabular}
\end{table}

\vspace{-0.2in}
\section{Self-Evolving Reinforcement Learning}
\label{app:rl}

\subsection{Why Self-Rewarding RL?}
\label{app:why_selfrewarding}

In this section, we discuss why standard reward-based reinforcement learning (RL) paradigms are not directly applicable to latent reasoning in UMMs, motivating the self-rewarding formulation adopted in \sname.

\noindent\textbf{Inapplicability of preference-based methods.} Direct Preference Optimization (DPO)~\citep{rafailov2023direct} and its diffusion variants such as Diffusion-DPO~\citep{wallace2024diffusion} and CaPO~\citep{lee2025calibrated} optimize the policy from paired preference data $(\tau^w, \tau^l)$, where $\tau^w$ is preferred over $\tau^l$. This paradigm presumes the existence of a meaningful preference order between trajectories. In our setting, however, each trajectory $\tau$ contains continuous latent steps $h(\tau)$ that lie outside any human-interpretable space, ruling out direct human annotation over the reasoning trace. External reward models for image generation~\citep{xu2023imagereward,kirstain2023pick,kim2024confidence} also operate on the rendered image $x$ rather than the latent chain, providing no signal that distinguishes between two latent traces that yield comparable images. A recent study~\citep{tong2025delving} further reports that DPO-based optimization in autoregressive image generation tends to overfit to in-domain preference distributions, limiting generalization beyond the curated preference set. These observations indicate that preference-based methods are structurally misaligned with the latent reasoning regime, which lacks both pairwise preference labels and a natural notion of pairwise dominance over hidden states.

\noindent\textbf{Limitations of critic-based methods.} Proximal Policy Optimization (PPO)~\citep{schulman2017ppo} sidesteps the preference assumption but introduces a separate value network that estimates the expected return at each position, which roughly doubles the trainable parameters and adds non-trivial training overhead at the scale of UMMs. More fundamentally, fitting an accurate critic over the latent reasoning trajectory requires a dense external supervision signal, which is precisely what is absent in our setting. Without ground-truth latent annotations, the critic must rely on an external image-quality estimator or a learned reward model, both of which inherit the same evaluation gap described above.

\noindent\textbf{Self-rewarding via information gain.} We point out two requirements that the optimization framework must satisfy. (1) The reward should be derivable from the model itself, since no external evaluator can faithfully assess latent reasoning traces. (2) The policy gradient estimator should avoid an additional value network to keep training tractable on top of the UMM backbone. In \sname, the information-gain reward satisfies the first requirement by reusing the frozen reference $\pi_{\text{ref}}$ obtained from Stage 1, scoring each trajectory through the change in target-image likelihood. Additionally, GRPO~\citep{shao2024deepseekmath} satisfies the second by replacing the value network with group-relative advantages estimated from multiple rollouts.

\subsection{Derivation of the Information-Gain Reward}
\label{app:info_gain_deriv}

We provide the full derivation of the Information-Gain reward $r_{\text{IG}}$ introduced in Section~\ref{sec:method:3.4}. Recall that the reward is defined as
\begin{equation}
r_{\text{IG}}(\tau \,;\, \mathbf{x}, \mathbf{c}) \triangleq \log p_{\pi_{\text{ref}}}(\mathbf{x} \mid \mathbf{c}, h(\tau)) - \log p_{\pi_{\text{ref}}}(\mathbf{x} \mid \mathbf{c}),
\label{eq:rig_def}
\end{equation}
where $h(\tau)$ denotes the sequence of latent hidden states produced along trajectory $\tau$, and $\mathbf{x}$ is the target image. Taking the expectation of Equation~\ref{eq:rig_def} under the joint distribution $(\mathbf{x}, \mathbf{c}, \tau)$ recovers the conditional mutual information,
\begin{equation}
\mathbb{E}\!\left[\, r_{\text{IG}} \,\right] = \mathbb{E}\!\left[\, \log \frac{p_{\pi_{\text{ref}}}(\mathbf{x} \mid \mathbf{c}, h(\tau))}{p_{\pi_{\text{ref}}}(\mathbf{x} \mid \mathbf{c})} \,\right] = I_{\pi_{\text{ref}}}\!\left( \mathbf{x} \,;\, h(\tau) \mid \mathbf{c} \right) \;\geq\; 0.
\label{eq:mi_lb}
\end{equation}
Maximizing $r_{\text{IG}}$ is therefore equivalent to maximizing the mutual information between the latent reasoning trace and the target image, with the bare context $c$ playing the role of the marginal baseline. The non-negativity in Equation~\ref{eq:mi_lb} follows from Jensen's inequality applied to the log-ratio, which guarantees that the expected reward is bounded below by zero regardless of the policy.

Since the generation expert of UMM does not provide closed-form access to the marginal density $p_{\pi_{\text{ref}}}(\mathbf{x} \mid \cdot)$, we approximate the log-likelihood through the conditional flow-matching objective. Following the rectified flow formulation of BAGEL~\citep{deng2025bagel}, the negative log-likelihood admits the bound
\begin{equation}
-\log p_{\pi_{\text{ref}}}(\mathbf{x} \mid \xi) \;=\; \mathcal{L}_{\text{flow}}^{\pi_{\text{ref}}}(\mathbf{x}, \xi) + C(\mathbf{x}),
\label{eq:flow_nll}
\end{equation}
where $\xi$ denotes the conditioning variable, $\mathcal{L}_{\text{flow}}^{\pi_{\text{ref}}}$ is the conditional flow-matching loss evaluated under $\pi_{\text{ref}}$, and $C(\mathbf{x})$ is a sample-wise constant that depends only on the target image and the noise schedule. Substituting Equation~\ref{eq:flow_nll} into Equation~\ref{eq:rig_def} for both terms yields
\begin{align}
r_{\text{IG}}(\tau \,;\, \mathbf{x}, \mathbf{c})
&= \big[\, -\mathcal{L}_{\text{flow}}^{\pi_{\text{ref}}}(\mathbf{x}, \mathbf{c}, h(\tau)) - C(\mathbf{x}) \,\big] - \big[\, -\mathcal{L}_{\text{flow}}^{\pi_{\text{ref}}}(\mathbf{x}, \mathbf{c}) - C(\mathbf{x}) \,\big] \nonumber \\
&= \mathcal{L}_{\text{flow}}^{\pi_{\text{ref}}}(\mathbf{x}, \mathbf{c}) \,-\, \mathcal{L}_{\text{flow}}^{\pi_{\text{ref}}}(\mathbf{x}, \mathbf{c}, h(\tau)).
\label{eq:rig_flow}
\end{align}
Crucially, the constant $C(\mathbf{x})$ cancels exactly when the two flow-matching losses are evaluated under the same noise sample $\eta \sim \mathcal{N}(0, I)$ and the same time step $t \sim \mathcal{U}[0, 1]$, since both terms inherit the identical perturbation. We therefore implement the reward as a paired evaluation under shared randomness, as detailed in Algorithm~\ref{alg:rl}. This shared-noise estimator yields an unbiased and low-variance approximation of Equation~\ref{eq:rig_def}, requiring no additional reward model and reusing the frozen $\pi_{\text{ref}}$ that is already available from Stage 1.

\begin{algorithm}[ht]
\caption{Self-Evolving Reinforcement Learning of \sname}
\label{alg:rl}
\begin{algorithmic}[1]
\Require Frozen reference $\pi_{\text{ref}} = \mathcal{M}_{\text{SFT}}$, dataset $\mathcal{D}$, hyperparameters $G$, $\epsilon_{\text{clip}}$, $\beta$, $\lambda_{\text{div}}$
\State Initialize $\pi_\theta \gets \pi_{\text{ref}}$
\For{each RL step}
    \State $\pi_{\theta_{\text{old}}} \gets \pi_\theta$
    \State Sample $(\mathbf{c}, \mathbf{x}) \sim \mathcal{D}$ and rollout $\{\tau_i\}_{i=1}^{G} \sim \pi_{\theta_{\text{old}}}(\cdot \mid \mathbf{c})$
    \For{$i = 1, \ldots, G$}
        \State $r_{\text{IG}}(\tau_i) \gets \mathcal{L}_{\text{flow}}^{\pi_{\text{ref}}}(\mathbf{x}, \mathbf{c}) - \mathcal{L}_{\text{flow}}^{\pi_{\text{ref}}}(\mathbf{x}, \mathbf{c}, h(\tau_i))$ \Comment{Shared noise}
        \State $R_i \gets r_{\text{IG}}(\tau_i) + \lambda_{\text{div}} \cdot r_{\text{div}}(\tau_i)$
    \EndFor
    \State $A_i \gets (R_i - \mu_R) / (\sigma_R + \varepsilon)$ for $i = 1, \ldots, G$
    \State Update $\theta$ by maximizing $\mathcal{J}_{\text{GRPO}}(\theta)$ \Comment{See Equation~\ref{eq:grpo}}
\EndFor
\State \Return $\mathcal{M}_{\text{\sname}} \gets \pi_\theta$
\end{algorithmic}
\end{algorithm}

\subsection{Policy Optimization}

We instantiate the policy gradient optimization in Equation~\ref{eq:rl_obj} using GRPO~\citep{shao2024deepseekmath}, which estimates advantages from group rollouts and removes the need for a separate value network. For each context $c$ sampled from $\mathcal{D}$, we draw $G$ trajectories $\{\tau_i\}_{i=1}^{G}$ from the old policy $\pi_{\theta_{\text{old}}}$, compute the total reward $R_i$ for each rollout, and normalize within the group to obtain the advantage $A_i$. The policy is then updated by maximizing the clipped surrogate objective
\begin{equation}
\mathcal{J}_{\text{GRPO}}(\theta) = \mathbb{E}\!\left[ \frac{1}{G} \sum_{i=1}^{G} \min\!\Big( \rho_i(\theta) A_i, \; \mathrm{clip}(\rho_i(\theta), 1 - \epsilon_{\text{clip}}, 1 + \epsilon_{\text{clip}}) A_i \Big) - \beta \, \mathrm{KL}\!\left[ \pi_\theta \,\|\, \pi_{\text{ref}} \right] \right],
\label{eq:grpo}
\end{equation}
where $\rho_i(\theta) = \frac{\pi_\theta(\tau_i \mid \mathbf{c})}{\pi_{\theta_{\text{old}}}(\tau_i \mid \mathbf{c})}$ is the importance ratio. The full training procedure is summarized in Algorithm~\ref{alg:rl}.

\section{Additional Experimental Results}
\label{app:additional}

\subsection{Detailed Analysis on Test-time Scaling}
\label{app:tts}

We provide an additional analysis on performance and computational cost incurred by latent reasoning at test-time. We measure the per-sample clock-time on a single NVIDIA A100 with batch size $1$, varying the number of images $N$ and latent steps $L$ from $1$ to $10$. We compare \sname against best-of-$N$ sampling on the base model BAGEL~\citep{deng2025bagel}, which represents the standard test-time scaling baseline that draws $N$ independent samples and selects the best one. We analyze clock time (seconds) of test-time scaling in Table~\ref{tab:clock_time}.

\begin{table}[ht]
\centering
\caption{\textbf{Clock-time analysis under test-time scaling.}}
\label{tab:clock_time}
\small
\vspace{0.5em}
\begin{tabular}{lcccccc}
\toprule
Method & $N{=}1$ & $N{=}2$ & $N{=}4$ & $N{=}6$ & $N{=}8$ & $N{=}10$ \\
\midrule
BAGEL~\citep{deng2025bagel} + Best-of-$N$ & 46.5 & 92.8 & 187.1 & 275.5 & 367.7 & 469.0 \\
\midrule
& $L{=}1$ & $L{=}2$ & $L{=}4$ & $L{=}6$ & $L{=}8$ & $L{=}10$ \\
\cmidrule{2-7}
\sname (ours) & 46.8 & 46.9 & 47.0 & 47.3 & 47.6 & 48.0 \\
\bottomrule
\end{tabular}
\end{table}

Note that \sname only requires single-pass while best-of-$N$ demands iterative sampling.
As shown in Table~\ref{tab:clock_time}, \sname incurs only a marginal overhead as the number of latent steps increases, requiring additional $1.2$ seconds at $L=10$ compared to $L=1$. Notably, combined with the results shown in \Cref{fig:2}, \sname outperforms best-of-$N$ on every scale with highly efficient computational cost, performing consistent test-time scaling behavior.
We attribute this efficiency to the fact that latent reasoning extends only a small portion of the forward pass through the understanding expert, whereas best-of-$N$ requires repeated invocation of the full generation expert. These results confirm that continuous latent reasoning offers a practically efficient route to test-time scaling for UMMs.

\subsection{Additional Ablation Study}
\label{app:ablation}

\noindent\textbf{Ablation study for latent diversity reward.} We investigate the role of the collapse-prevention reward $r_{\text{div}}$ introduced in Equation~\ref{eq:total_reward}. Without $r_{\text{div}}$, we observe that the policy quickly collapses to trivial trajectories where all $N$ latent steps converge to nearly identical hidden states, eliminating the benefit of multi-step reasoning. To quantify this phenomenon, we report the average pairwise cosine similarity among latent steps along with task performance under varying $\lambda_{\text{div}}$ in Table~\ref{tab:diversity_ablation}.

\begin{table}[h]
\centering
\caption{\textbf{Effect of the latent diversity reward.} We report the average pairwise cosine similarity (Cos. Sim.) among latent steps within a trajectory, and accuracy on GenEval, OneIG-Bench, and MIRA.}
\label{tab:diversity_ablation}
\small
\vspace{0.5em}
\begin{tabular}{lcccc}
\toprule
$\lambda_{\text{div}}$ & Cos. Sim. & GenEval $\uparrow$ & OneIG-Bench $\uparrow$ & MIRA $\uparrow$ \\
\midrule
$0.0$ & 0.97 & 0.83 & 0.81 & 9.8 \\
$0.05$ & 0.87 & 0.85 & 0.84 & 10.9 \\
\rowcolor{RowHighlight}$0.1$ (ours) & 0.79 & 0.88 & 0.87 & 11.7 \\
$0.2$ & 0.71 & 0.85 & 0.85 & 11.4 \\
$0.5$ & 0.65 & 0.82 & 0.82 & 10.6 \\
\bottomrule
\end{tabular}
\end{table}

As shown in Table~\ref{tab:diversity_ablation}, removing the diversity reward ($\lambda_{\text{div}}{=}0$) leads to a pairwise cosine similarity of $0.97$ across latent steps, indicating severe collapse where every step encodes nearly redundant information. This collapse degrades all three benchmarks, with MIRA dropping from $11.7$ to $9.8$. Setting $\lambda_{\text{div}}{=}0.1$ yields the best trade-off. These results validate that the diversity regularizer is essential for preventing latent collapse and confirm that the chosen weight balances exploration of distinct latent directions with exploitation of informative ones.

\clearpage

\subsection{Qualitative Results}

\null
\vspace{1.2in}
\begin{figure}[H]
    \centering
    \includegraphics[width=\linewidth]{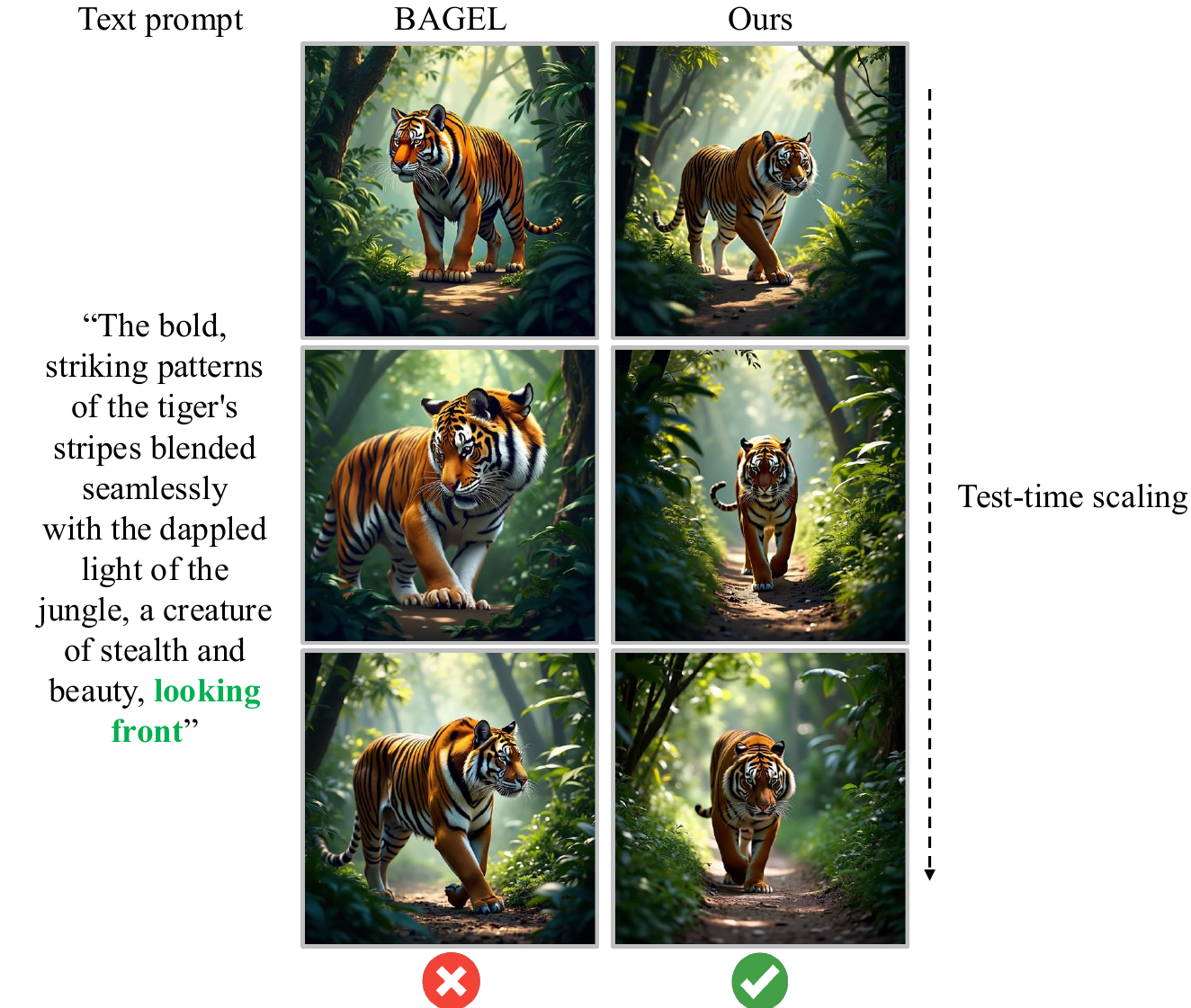}
    \caption{
        \textbf{Qualitative comparison during test-time.} We compare BAGEL~\citep{deng2025bagel} with best-of-$N$ sampling to \sname on test-time scaling with $L$ latent steps on text-to-image generation task. First row presents $N=L=1$, second row presents $N=L=4$, and third row presents $N=L=10$.
    }
    \label{fig:test_time_scaling}
\end{figure}

\clearpage

\null
\vspace{1.2in}
\begin{figure}[H]
    \centering
    \includegraphics[width=\linewidth]{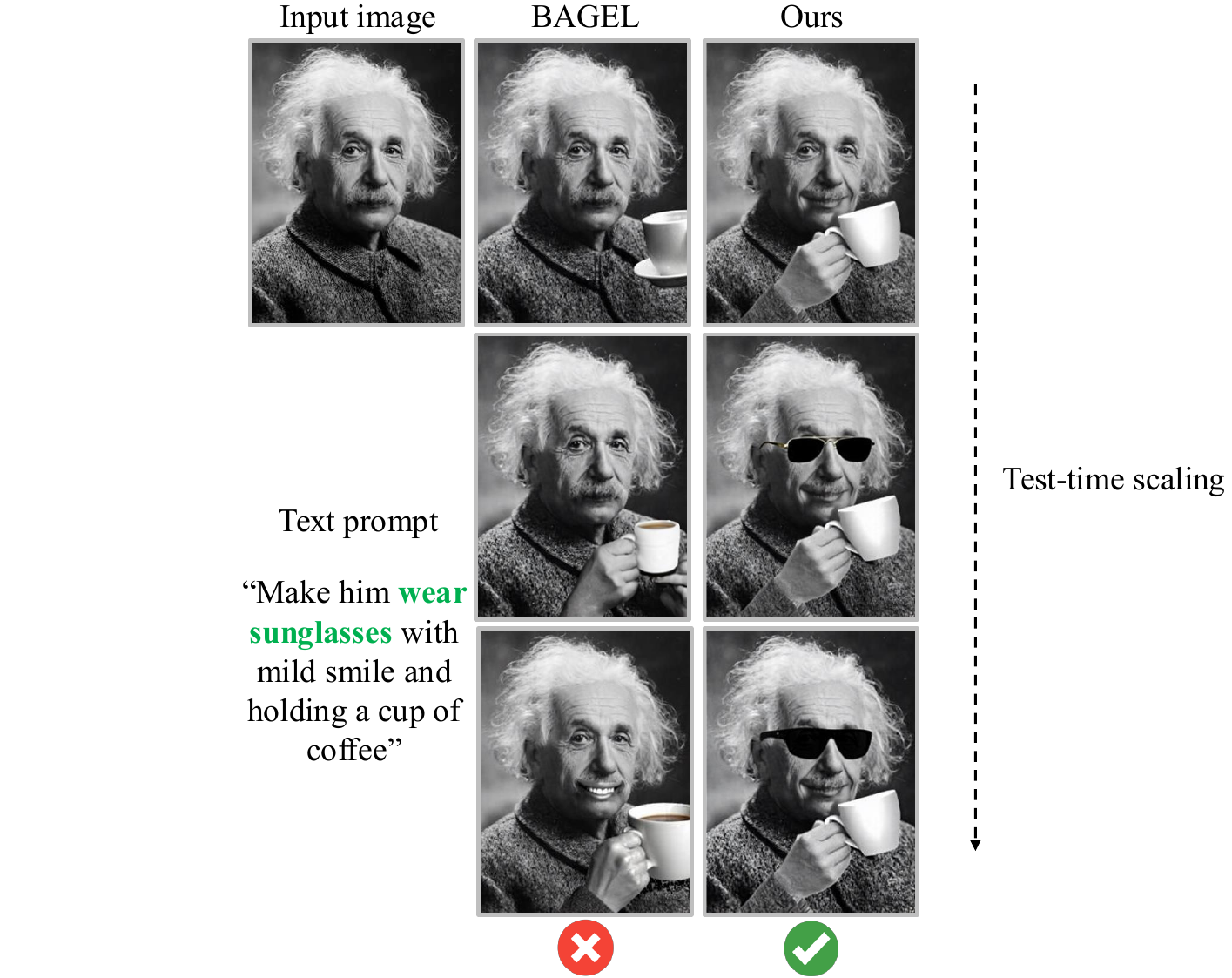}
    \caption{
        \textbf{Qualitative comparison during test-time.} We compare BAGEL~\citep{deng2025bagel} with best-of-$N$ sampling to \sname on test-time scaling with $L$ latent steps on image editing task. First row presents $N=L=1$, second row presents $N=L=4$, and third row presents $N=L=10$.
    }
    \label{fig:test_time_scaling_edit}
\end{figure}

\clearpage

\section{Limitations and Future Work}
\label{sec:limitations}
While \sname demonstrates strong empirical gains and clear test-time scaling on various generation tasks, several directions remain open for future exploration. First, our framework is currently instantiated on UMMs whose image module is a flow-matching head, which generates the image in continuous space. Extending the same principle to UMMs with discrete space generation or other generative heads is a natural next step. Second, although \sname enables effective test-time scaling, the interleaved latent reasoning trajectory introduces additional computation at inference, leading to latency overhead compared to the base UMM. Reducing this overhead through efficient decoding strategy, distillation, or adaptive latent allocation is a promising research direction. We leave these extensions to future work.

\end{document}